# BioSentVec: creating sentence embeddings for biomedical texts

Qingyu Chen[†]    Yifan Peng[†]    Zhiyong Lu[*]

National Center for Biotechnology Information (NCBI), National Library of Medicine (NLM), National Institutes of Health (NIH)
8600 Rockville Pike, Bethesda, MD 20894, USA
{qingyu.chen, yifan.peng, zhiyong.lu}@nih.gov
[†]Equal contributions. [*]To whom correspondence should be addressed.

*Abstract*— Sentence embeddings have become an essential part of today's natural language processing (NLP) systems, especially together advanced deep learning methods. Although pre-trained sentence encoders are available in the general domain, none exists for biomedical texts to date. In this work, we introduce BioSentVec: the first open set of sentence embeddings trained with over 30 million documents from both scholarly articles in PubMed and clinical notes in the MIMIC-III Clinical Database. We evaluate BioSentVec embeddings in two sentence pair similarity tasks in different biomedical text genres. Our benchmarking results demonstrate that the BioSentVec embeddings can better capture sentence semantics compared to the other competitive alternatives and achieve state-of-the-art performance in both tasks. We expect BioSentVec to facilitate the research and development in biomedical text mining and to complement the existing resources in biomedical word embeddings. The embeddings are publicly available at https://github.com/ncbi-nlp/BioSentVec.

## I. INTRODUCTION

Capturing sentence semantics plays a critical role in biomedical and clinical text mining research. Traditional methods that rely on bag-of-words may not model such information accurately due to natural language ambiguity. For instance, different sentences can be used to describe similar findings (e.g., 'It has recently been shown that Craf is essential for Kras G12D-induced NSCLC.' versus 'It has recently become evident that Craf is essential for the onset of Kras-driven non-small cell lung cancer.'[1]. In response, embedding-based approaches have shown promising results recently as the semantic is represented by high dimensional vectors regardless whether the same set of words are used. Such vector-based representations are commonly learnt from large text corpora [2, 3] have becoming increasingly important in today's text mining research, especially when used as input in advanced deep learning (DL) techniques [4, 5].

Representative sentence embeddings are doc2vec [6], Universal Sentence Encoder [7], and sent2vec [8]. However, to the best of our knowledge, there is no publicly available sentence embeddings in biomedicine and clinical domains, in spite of many related use cases and sentence-based applications, such as finding relevant sentences for information retrieval [1, 9], clinical sentence similarity [10], biomedical sentence classification [11], or biomedical question answering [12]. As a result, researchers would need either train sentence embeddings on their own from scratch (a data and time-intensive process, together with selection of best model parameters), derive them from individual word embeddings (loss of information about the entire sentence), or use pretrained sentence embeddings from the general domain (may suffer from the out-of-domain issue). Due to these problems, suboptimal performance may be obtained.

Table I. Corpora for training embeddings.

| Corpus | Documents | Sentences | Tokens |
|---|---|---|---|
| PubMed | 28,714,373 | 181,634,210 | 4,354,171,148 |
| MIMIC-III | 2,083,180 | 41,674,775 | 539,006,967 |

To facilitate text mining research in biomedicine, we propose BioSentVec, a pre-trained sentence embeddings for readily generating sentence vectors given any arbitrary sentences as inputs. Specifically, BioSentVec is created by applying sent2vec, an advanced unsupervised model, to both biological and clinical texts at a large scale. BioSentVec is evaluated on two independent tasks: sentence similarity and multi-label text classification, and is compared to the current state-of-the-art methods.

## II. METHODS AND MATERIALS

To maximize the robustness and generalizability of BioSentVec on different text genres in biomedicine and clinical domains, BioSentVec embeddings are trained using both PubMed and the clinical notes from MIMIC-III Clinical Database [13]. Collectively, they consist of over 30 million documents, ~223 million sentences, and ~5 billion tokens. Table I summarizes the detailed statistics of the corpora.

Both PubMed and MIMIC-III texts were sentence-split and tokenized using NLTK [14]. We then trained BioSentVec using sent2vec [8]. It adapts the Continuous Bag-of-Words model - known for training word embeddings – at the sentence level, and extends the model by using n-grams of sentences. By far, it achieves the state-of-the-art performance in a range of text mining tasks in the general domain [8]. Based on a set of experiments with different parameter settings (the vector dimension, window size and negative samples), we empirically obtained 700-dimensional vectors by applying the bigram model of sent2vec with a window size of 30 and negative examples of 10, as BioSentVec was seemingly robust in this set-up. Full-text articles in PMC were also attempted but no overall performance gain was observed, which is consistent with findings in the past [4]. We evaluate the effectiveness of BioSentVec in the following two tasks.

### A. Task 1: Biological and clinical sentence similarity

We evaluate BioSentVec for the task of finding similar sentences on two separate datasets. BIOSSES consists of 100 sentence pairs from PubMed articles annotated by 5 curators [1]. MedSTS consists of 1,068 sentence pairs annotated by 2 curators from clinical notes where 750 and 318 pairs are used for training and testing, respectively [10]. The similarity of sentence pairs is annotated in terms of five categories from 0 (not similarity) to 5 (very similar), and the annotation scores on the same pair by different curators are averaged to be the final similarity score for that pair, i.e., the final label a is continuous number ranging from 0 to 5. The detailed description of the annotation can be found in [1, 10].

Both unsupervised and supervised methods have been attempted in the past on these two datasets. In the unsupervised fashion, previous best-performing approaches used doc2vec and Levenshtein Distance for BIOSSES and MEDSTS, respectively [1, 10]. In addition, we propose several alternatives: sentence embeddings derived by averaging word embeddings (pre-trained in [4]) or directly from pretrained sentence encoders (BioSentVec or Universal Sentence Encoder).

In the supervised setting, previous methods made use of ensemble learning that aggregates a range of human engineered features, including the use of embeddings [1, 15]. In comparison, we propose a straightforward 5-layer deep learning model, which takes two sentence vectors (generated by BioSentVec) as inputs. The first layer concatenates the two vectors, their absolute differences, multiplications and the dot product. It is followed by three fully-connected layers, having 256, 128, 64 hidden units, respectively. We used ReLU as the activation function and selected a dropout rate at 0.5. The final layer is the prediction layer, which outputs the predicted similarity for a sentence pair. In terms of hyperparameters, we used Xavier normal initializer for weight initialization. We set the values of bias as a constant of 0.01, and applied L2 regularization. When training the model, we used SGD as its optimizer with a learning rate of 0.001. We chose mean squared error as the loss function, and set the batch size to be 8. The model was trained via 1500 epochs and the one with the lowest loss on the validation set was saved.

The preprocessing and the evaluation are the same for both unsupervised and supervised methods. For preprocessing, stop words and punctuations are removed in all these sentences. For evaluation, pearson correlation coefficient is used to compare the algorithm results with the gold standard according to the literature. We also follow the exact evaluation procedure accordingly. For BIOSSES, we use 10-fold stratified cross-validation as described in [1]. For MEDSTS, we report the results on the official test set.

### B. Task 2: Biomedical multi-label text classification

We also evaluate BioSentVect for the task of multi-label text classification on the Hallmarks of Cancer corpus (https://www.cl.cam.ac.uk/~sb895/HoC.html) [16, 17]. This public dataset consists of 14,919 sentences, each of which was manually annotated with 10 currently known hallmarks of cancer, i.e., each sentence has one or more labels from these 10 hallmarks. We randomly chose 60% of the Hallmarks corpus for training, 20% for development, and 20% for test.

To classify each sentence, we developed a CNN-based model that takes as input both word and the sentence embeddings and outputs a 10-element probability vector corresponding to the 10 cancer hallmarks (see Fig. 1). We chose the CNN model because CNNs stand out as the most popular model both in terms of computational complexity and performance amongst different deep learning architectures, and have competitive performance on tasks related with short text [19].

This model consists of three layers: an embedding layer, a convolution layer (window size of 3 and 100 filters), and three fully-connected layers (having 256, 128 and 10 hidden units, respectively). In the embedding layer, each word in a sentence is first represented by concatenating its own embedding, part-

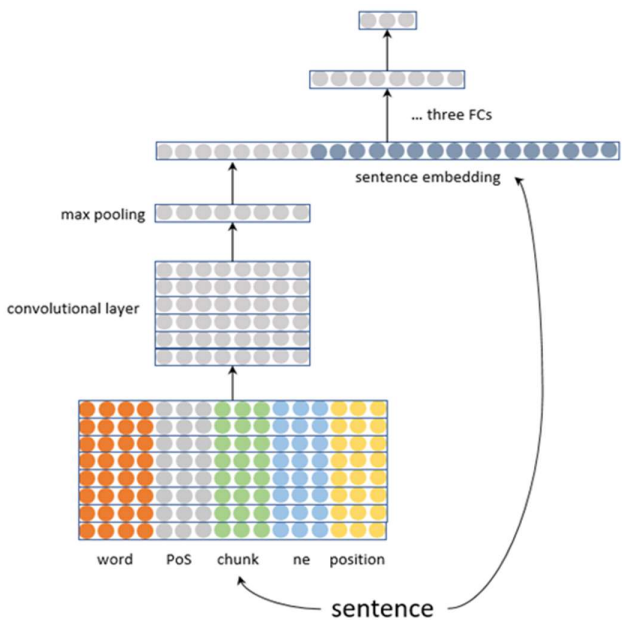

Fig. 1. Framework of the CNN model for multi-label text classification.

Table II. Sentence similarity evaluation results for Task 1.

|  | BIOSSES | MEDSTS |
|---|---|---|
| **Unsupervised** | | |
| Doc2vec [1]* | 0.787 | - |
| Levenshtein Distance[10]*† | - | 0.680 |
| Averaged word embeddings[4] | 0.694 | 0.747 |
| Universal Sentence Encoder[7] | 0.345 | 0.714 |
| BioSentVec (PubMed) | **0.817** | 0.750 |
| BioSentVec (MIMIC-III) | 0.350 | 0.759 |
| BioSentVec (PubMed + MIMIC-III) | 0.795 | **0.767** |
| **Supervised** | | |
| Linear Regression[1]* | 0.836 | - |
| Random Forest[18]* | - | 0.818 |
| Deep learning + Avg. word embeddings[4] | 0.703 | 0.784 |
| Deep learning + Universal Sentence Encoder[7] | 0.401 | 0.774 |
| Deep learning + BioSentVec (PubMed) | 0.824 | 0.819 |
| Deep learning + BioSentVec (MIMIC-III) | 0.353 | 0.805 |
| Deep learning + BioSentVec (PubMed + MIMIC-III) | **0.848** | **0.836** |

\* The previous best results in terms of single models in literature.
† No embeddings were used.

# BioSentVec Tutorial

This tutorial provides a fundemental introduction to our BioSentVec models. It illustrates (1) how to load the model, (2) an example function to preprocess sentences, (3) an example application that uses the model and (4) further resources for using the model more broadly.

## 1. Prerequisites

Please download BioSentVec model and install all the related python libraries

```python
import sent2vec
from nltk import word_tokenize
from nltk.corpus import stopwords
from string import punctuation
from scipy.spatial import distance
```

## 2. Load BioSentVec model

Please specify the location of the BioSentVec model to model_path. It may take a while to load the model at the first time.

```python
model_path = YOUR_MODEL_LOCATION
model = sent2vec.Sent2vecModel()
try:
    model.load_model(model_path)
except Exception as e:
    print(e)
print('model successfully loaded')
```

model successfully loaded

## 3. Preprocess sentences

Fig. 2. An executable Jupyter Notebook provides examples for applying BioSentVec models.

Table III. Multi-label classification evaluation results for Task 2.

| Method | Prec | Recall | F1 |
|---|---|---|---|
| Deep learning | 0.546 | 0.503 | 0.524 |
| Deep learning + Avg. word embeddings[4] | 0.615 | 0.587 | 0.601 |
| Deep learning + Universal Sentence Encoder[7] | 0.648 | 0.626 | 0.637 |
| Deep learning + BioSentVec (MIMIC-III) | 0.652 | 0.623 | 0.637 |
| Deep learning + BioSentVec (PubMed) | 0.669 | 0.645 | 0.657 |
| Deep learning + BioSentVec (PubMed + MIMIC-III) | **0.677** | **0.654** | **0.665** |

of-speech, and chunk features, among others. The part-of-speech and chunk features are encoded using a one-hot schema, following the previous study [20]. Next, we apply convolution to input sentences to first obtain local features, and then a global feature through max pooling. The resulting feature vector is then concatenated with the embeddings directly generated based on sentences, followed by three fully-connected layers for sentence classification. We trained the model using Adam optimizer with a batch size of 64 and a learning rate of 7e-4. The training was stopped when the loss on the validation set does not decrease for 10 epochs.

We evaluate a total of six models with regards to if and which sentence embeddings were used: a CNN model without sentence embeddings, and five models with different sentence embeddings as in Task 1. Exampled-based Precision, Recall and F1-score were used as evaluation metrics for multi-label classifications [21].

## III. RESULTS AND DISCUSSIONS

For the sentence similarity task, Table II shows the evaluation results. It shows that the highest performance was obtained based on the proposed BioSentVec embeddings in both supervised and unsupervised methods, suggesting BioSentVec can capture sentence meanings consistently better than other approaches such as averaged word vectors or sentence vectors trained from the general domain.

For example, a single end-to-end DL model is able to achieve the best results without the need of a complex ensemble approach. BioSentVec also appear to be robust and generalizable on different text genres in biomedicine. Furthermore, the results in Table I show that BioSentVec trained from both PubMed and clinical notes generally have better performance than from a single source (with one exception). It also shows that BioSentVec trained from clinical notes only is not sufficient when used in applications dealing with PubMed articles.

For the second multi-label text classification task, our results are summarized in Table III. Similar to the previous task, BioSentVec trained on both PubMed and MIMIC-III achieved the best results amongst all the methods. Noticeably, the deep learning approach in this specific task only achieves ~50% F1-score; in contrast, using BioSentVec features doubled the performance.

While we evaluate a range of baseline models including the current state-of-the-art models, a limitation of the study is that it does not evaluate sentence embeddings trained from the biomedical domain. This is because, to our best knowledge, there is no publicly available sentence embedding trained from the biomedical domain. Instead, we evaluate using word embeddings trained from biomedical corpora as alternatives.

Also importantly, BioSentVec has already been used in practice. An adapted version of BioSentVec has been applied in LitSense [22]: a web server for searching across ~29 million PubMed abstracts and ~3 million PMC full-text articles at sentence level. It shows that combining traditional term matching based methods and BioSentVec can significantly improve the search effectiveness of sentence retrieval.

To facilities the convenient use of BioSentVec, we provide a Jupyter Notebook which contains code examples for applying BioSentVec models (see Fig. 2). It summarizes the fundamental usage steps, and provides example functions for preprocessing sentences, along with a simple example of computing sentence similarities using BioSentVec.

## IV. CONCLUSION

In summary, we introduced a new set of sentence embeddings pretrained on two different corpora in biomedicine and clinical domains and demonstrated its superior performance in benchmarking. All the embeddings trained with different sources, together with our deep learning models, are made publicly available. We hope BioSentVec can facilitate the development of deep learning models and text mining applications in biomedical research. In the future, we will conduct a more in-depth evaluation to quantify the effectiveness of sentence embeddings and language models in a range of real-world biomedical applications.

ACKNOWLEDGMENT

This research was supported by the Intramural Research Program of the NIH, National Library of Medicine. The authors thank Dr. Yijia Zhang and Dr. Alexis Allot for helpful discussion. We are grateful to the authors of sent2vec, BIOSSES, and MedSTS for making their software and data publicly available.